\pdfoutput=1

\documentclass[11pt, letterpaper]{berkeley}

\usepackage{times}
\usepackage{latexsym}

\usepackage[utf8]{inputenc}

\usepackage{microtype}

\usepackage{inconsolata}

\usepackage{graphicx}

\usepackage{amsmath}

\title{\texttt{PERSONA}: A Reproducible Testbed for Pluralistic Alignment}

\reportnumber{} 

\author[1]{Louis Castricato*}
\author[1]{Nathan Lile*}
\author[2]{Rafael Rafailov}
\author[2]{Jan-Philipp Fränken}
\author[2]{Chelsea Finn}

\affil[1]{SynthLabs.ai\footnote{Correspondence to team@synthlabs.ai}}
\affil[2]{Stanford University}

\begin{abstract}
The rapid advancement of language models (LMs) necessitates robust alignment with diverse user values. However, current preference optimization approaches often fail to capture the plurality of user opinions, instead reinforcing majority viewpoints and marginalizing minority perspectives. We introduce PERSONA, a reproducible test bed designed to evaluate and improve pluralistic alignment of LMs. We procedurally generate diverse user profiles from US census data, resulting in 1,586 synthetic personas with varied demographic and idiosyncratic attributes.

We then generate a large-scale evaluation dataset containing 3,868 prompts and 317,200 feedback pairs obtained from our synthetic personas. Leveraging this dataset, we systematically evaluate LM capabilities in role-playing diverse users, verified through human judges, and the establishment of both a benchmark, PERSONA Bench, for pluralistic alignment approaches as well as an extensive dataset to create new and future benchmarks.

The full dataset and benchmarks are available here \url{https://www.synthlabs.ai/research/persona}.

\end{abstract}

\begin{document}

\maketitle

\section{Introduction}
While reinforcement learning from human feedback (RLHF) approaches have been widely successful in creating helpful language model assistants \cite{ouyang2022training, geminiteam2024gemini, llama2024meta}, these algorithmic methods inherently instill opinions and values within the model based on the preferences expressed by the feedback providers. Recent works \cite{santurkar2023whose, lee2023large} have shown that widely used models do not in fact reflect the full diversity of demographic preferences---including on important topics---such as political biases \cite{rettenberger2024assessing, bang2024measuring}. These effects stem from both the opinions inherent within the user feedback data, but also the alignment algorithms used to train these models. Currently used practical methods do not take into account the plurality of users and difference of opinion, but instead work under the framework of a ``representative'' user, which may contribute to reinforcing majority opinions. 

Several recent studies have attempted to address this issue by developing algorithms that are specifically designed to account for the distributional nature of user values \cite{zhao2023group, chakraborty2024maxmin, siththaranjan2024distributional, ramesh2024group}. These approaches aim to align language models with the diverse preferences and opinions of different user groups, rather than focusing on a single ``representative'' user. However, significant challenges remain in achieving true pluralistic alignment \cite{sorensen2024roadmap}. Here, recent work has suggested it is not possible to simultaneously satisfy all group preferences with a single model \cite{chakraborty2024maxmin}, which may put into question the entire RLHF formulation. Going beyond distributional or group-level preferences, there is additional significant idiosyncratic variability in individual user values. In fact, these idiosyncratic values can be an even bigger driver of preferences than group-level attributes \cite{hwang2023aligning}. When properly aligned to individuals, generative models present opportunities to create uniquely bespoke interfaces, experiences and applications on a per user basis, which has recently driven significant research efforts into personalized alignment approaches \cite{jang2023personalized,li2024personalized, sun2024personadb}. Moreover, there have been a number of developments focused on active learning \cite{ji2024reinforcement, mehta2023sample, muldrew2024active, zhang2024selfexploring} and preference elicitation \cite{li2023eliciting, piriyakulkij2023active, andukuri2024stargate}, which aim to teach models to effectively learn about users from interactions. \textbf{However, one major challenge for the development and deployment of such approaches is evaluation}. 

Despite the significant amount of prior works and the practical importance of these problems, current test environments are still quite limited due to the challenging nature of not only collecting diverse and personalized preferences but evaluating the resulting models under those same users. Prior works \cite{santurkar2023opinions, zhao2023group, durmus2023towards, hwang2023aligning} have established opinion polls and population surveys as benchmark. However, these usually consist of multi-choice questions and do not reflect the actual use case of LMs. Moreover, accurately predicting user choices is not necessarily correlated to the LM's ability to generate responses that align with them \cite{rafailov2024scaling}. In addition such polls usually only cover group-level characteristics of the surveyed population and rarely contain detailed information about specific users, limiting their usefulness for personalization applications. One major recent development is the PRISM dataset \cite{kirk2024prism}, which collects preferences on actual LM-generated content from a wide arrange of global respondents on diverse and potentially controversial topics, with significant disagreement. While this effort provides good coverage for the problems discussed before, evaluation remains challenging as data is collected from real human respondents and thus algorithms and models cannot be evaluated in the same setting. 

\textbf{In this work we seek to address this evaluation issue through synthetic personas} \cite{xu2024character, joshi2024personas, chen2024persona}: We model personas with realistic user profiles including detailed demographic information and varied idiosyncratic individual background, which we use to set-up role-playing LMs. Following demographic surveys, user marketing profiles and prior work we create a broad representative demographic of 1,586 personas, which we use to generate diverse feedback on a number of value-laden, diverse, and controversial topics sampled from \cite{kirk2024prism}. Overall, we make the following \textbf{contributions}: First we systematically evaluate current LM capability to role-play as diverse users and verify our results with real human subjects study. We then create a benchmark of \textbf{1,586} synthetic personas as well as a large scale preference dataset with \textbf{3,868} prompts and \textbf{317,200} pairs of diverse feedback as provided by individual personas split into several datasets. Our data and evaluation framework can be used as (1) a test-bed, (2) a development environment, a (3) reproducible evaluation of pluralistic alignment approaches, (4) as personalization of LMs, and (5) for preference elicitation.

\begin{figure*}
    \centering
    \includegraphics[scale=0.3]{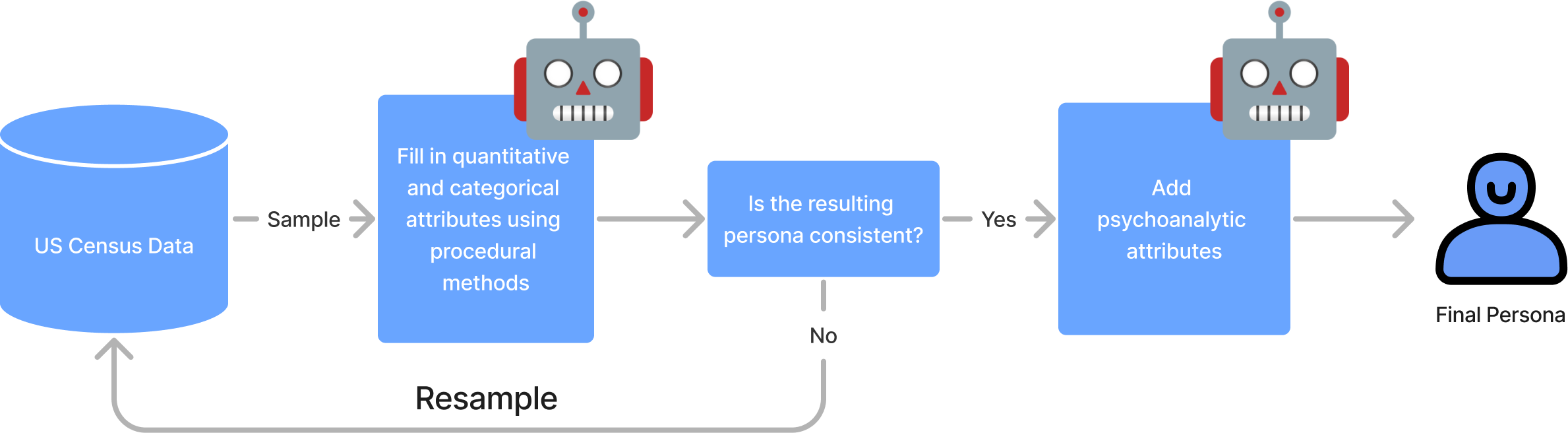}
    \caption{Procedure for generating personas. The above is a flow graph outlining the generation of a single persona. An exact example for this generation process can be found in the appendix. First, we sample a subset of US census data and query a language model to see if the resulting persona is self consistent. If it isn't, we resample. Next, we use procedural methods to fill in missing components of the census data. The list of procedural methods can be found in the appendix. Finally, we use a language model to fill in open ended psychoanalytic attributes.}
    \label{fig:persona_generation}
\end{figure*}

\section{Related Work}\label{sec:related_works}

\textbf{Challenges in Pluralistic Alignment.} While LMs are trained on data authored by billions of internet users, this involvement is passive, and pre-training datasets over-represent certain demographics \cite{wang2023not}, which can marginalize minority communities \cite{blodgett2020language, hershcovich2022challenges}. Moreover, while the RLHF process is paramount on instilling values within an LM it relies on even smaller pools of labellers \cite{sorensen2024roadmap}. This can manifest in misalignment between LM outputs and the views of diverse demographics including on major political and demographical divides \cite{santurkar2023whose, durmus2023towards, liu2024generation}. Moreover, \cite{chakraborty2024maxmin} theoretically show that a single model cannot simultaneously align with diverse groups holding conflicting opinions, calling into question the main objective of RLHF tuning \cite{sorensen2024roadmap}. Various approaches have been proposed to address these challenges, such as learning multiple reward models \cite{chakraborty2024maxmin, chidambaram2024direct}, latent variable models \cite{siththaranjan2024distributional, chidambaram2024direct}, preference elicitation \cite{andukuri2024star, li2023eliciting}, and few-shot alignment \cite{zhao2023group, shaikh2024show}. However, despite these advancements, pluralistic alignment remains a challenging problem.

\textbf{Evaluation of Pluralistic Alignment.} Pluralistic alignment approaches necessitates assessing how well methods actually align LMs with the range of human opinions captured in datasets. Datasets like OpinionQA \cite{santurkar2023whose}, GlobalOpinionQA \cite{durmus2023towards}, and opinion polls \cite{hwang2023aligning} have been widely used, but they only consist of multiple-choice questions and do not reflect realistic use cases of LMs. Other works have also used small-scale synthetic experiments or simple bimodal datasets, such as HH-RLHF \cite{bai2022constitutional}, which is not representative of real world distributional views.
The PRISM dataset \cite{kirk2024prism} makes progress in this direction by collecting a diverse set of open-ended conversations from a wide global population. However, it relies on human participants to provide feedback to LMs, which prevents scalable evaluation algorithms and models under the same distribution. 

\textbf{Role-Playing Language Agents.} Recent works have shown that LMs can emulate diverse personas and traits by leveraging prompts \cite{li2023chatharuhi, franken2023social, chen2024persona, xu2024character}, inherent knowledge \cite{shao2023character, lu2024large}, and finetuning \cite{park2023generative, franken2024self}. Carefully designed role-playing scenarios with such agents could provide the rich, controllable test-bed needed to evaluate alignment approaches without human participants.

\begin{figure*}
    \centering
    \includegraphics[width=0.975\textwidth]{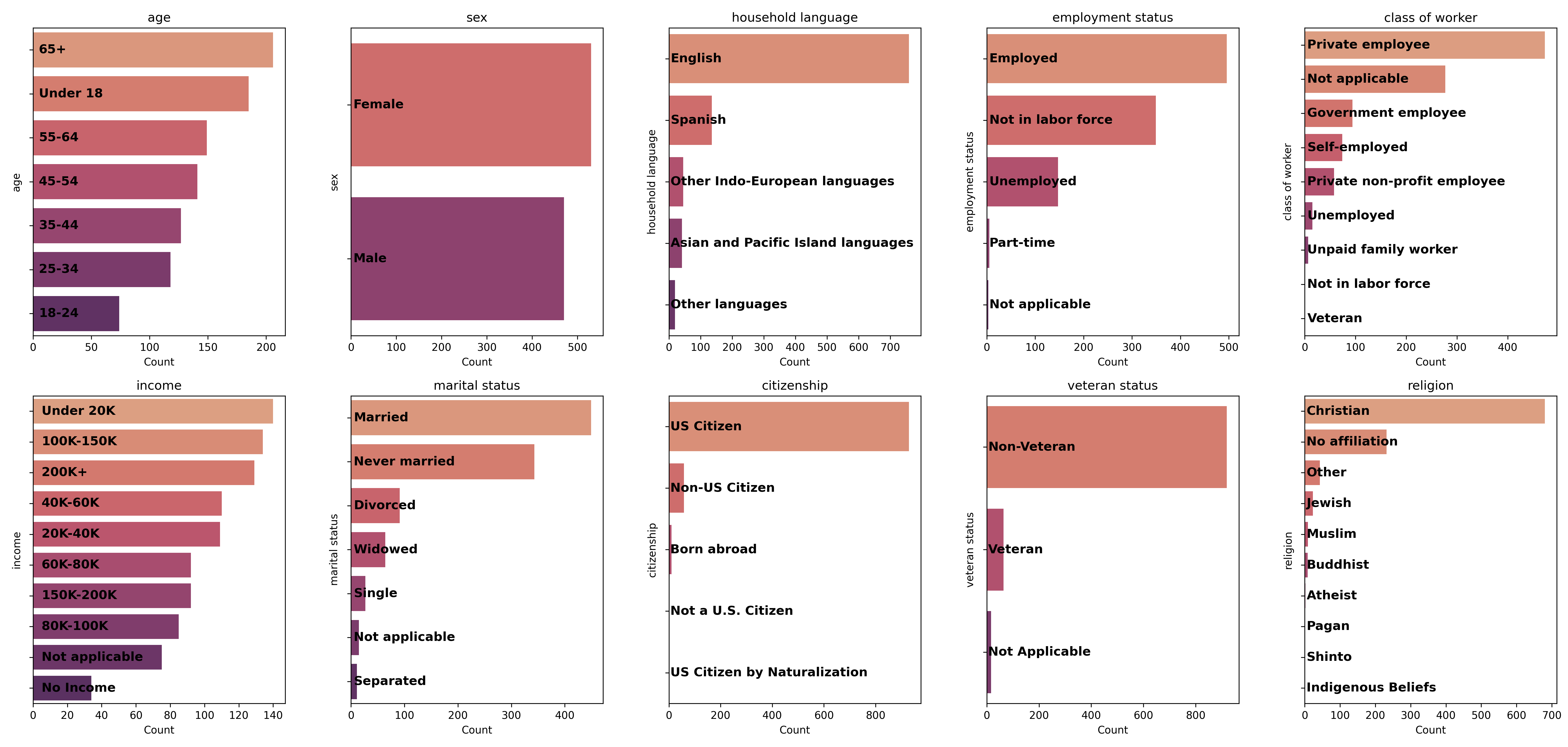}
    \caption{Histograms of group statistics of our demographic of synthetic personas.}
    \label{fig:persona_demographics}
\end{figure*}



\section{\texttt{PERSONA}: A Testbed for Pluralistic Alignment}
In this section, we outline the construction of our demographic of personas and the subsequent preference data generation process.

\subsection{Creating a Demographic of Personas}
Our full persona-generation pipeline is shown in \autoref{fig:persona_generation}. Within the taxonomy of \cite{chen2024persona}, our synthetic personas have a \textbf{demographic} and \textbf{individual} component. To construct demographic personas that accurately reflect the challenges of pluralistic alignment in a realistic setting, we construct a set of personas with demographics closely following the US population. This is challenging since standard US census data provides aggregate information across attributes but limited intersectional data and no personal characteristics. In contrast, the Census Bureau's American Community Survey (ACS) Public Use Microdata Sample (PUMS) files contain survey results from real people, making them more suitable for our purpose. Our dataset construction consists of several parts: (1) sampling from the PUMS files, (2) enriching each profile with additional statistically accurate psychodemographic data, (3) using language models to further enrich a small subset of fields, and (4) resolving inconsistencies (or pruning) with GPT-4.

We directly sample a subset of attributes from the PUMS files that cannot easily be self-inconsistent, such as someone under 18 making hundreds of thousands of dollars a year. Based on the selected characteristics, we procedurally create a demographic user profile and query GPT-4 to further filter out inconsistent ones, removing approximately 8.5\% of configurations. Moreover, we used the probabilities of the Big Five personality characteristics (neuroticism, openness, conscientiousness, agreeableness, and extraversion) from the Big Five Inventory-2 (BFI-2) developed by \cite{soto2017next} to procedurally generate five factor model personality profiles while additional core values, quirks, and mannerisms were sampled from a hand-curated set (see Appendix). Prior literature from marketing and business emphasizes the importance of psychoanalytic attributes on personal decision-making, so we further include such characteristics in our persona construction during the second generation stage \cite{8400224}

We noticed that procedurally generating idiosyncratic parts of the personas proved challenging, due to intersectionality effects and the open-ended nature of the problem. In our approach we broke these attributes into a number of high level categories such as "Lifestyle", "Personality", etc.. (the full list with all categories is included in \ref{sec:persona_attributes}). We further selected a number of categories per persona in order to guarantee diverse coverage end and prompted GPT-4 with these to create the final open-ended persona profile. For an example of complete profiles, consult the Appendix \ref{sec:example_personas}.

The distributional statistics of our final demographic of synthetic personas and their comparisons to the overall US census are presented in Fig. \ref{fig:persona_demographics}.

\subsection{Preference Dataset Construction}
\begin{figure*}
    \centering
    \includegraphics[width=0.975\textwidth]{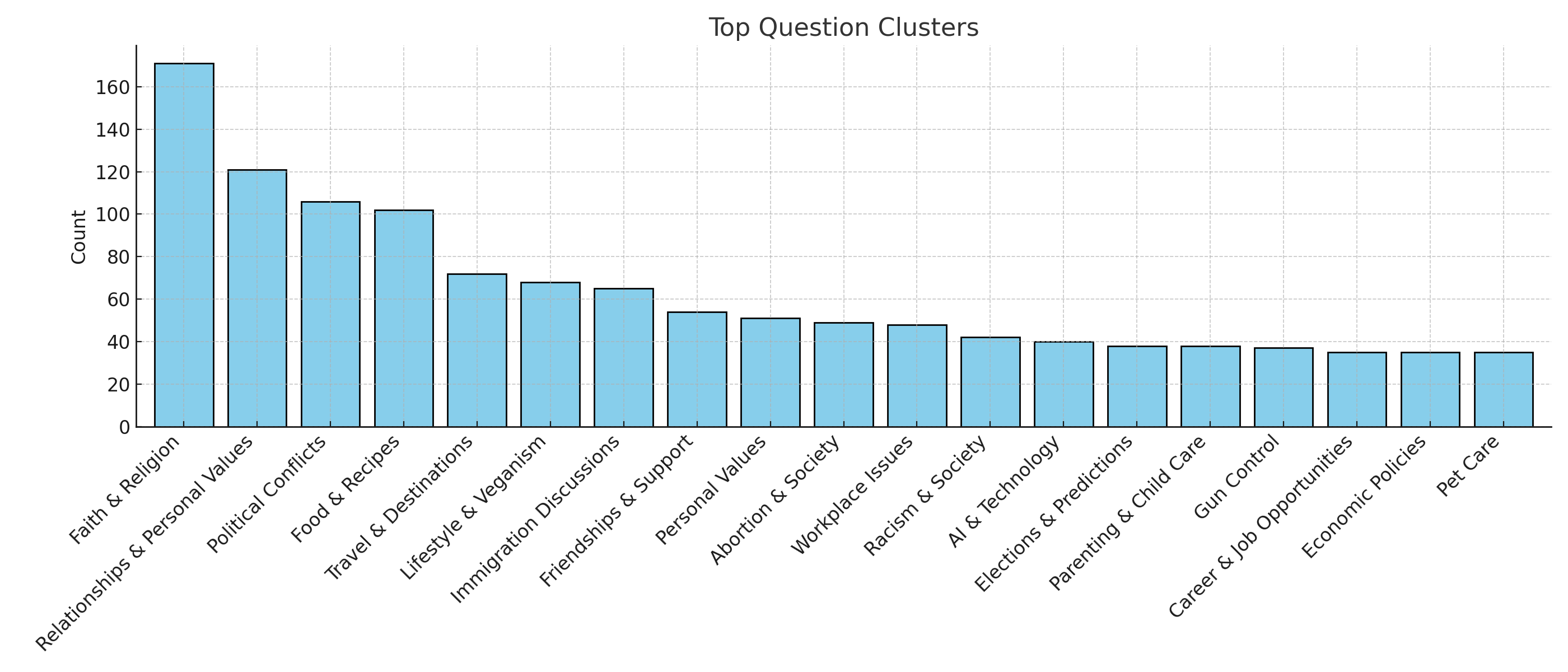}
    \caption{Distribution of prompt topics in the \texttt{Persona} dataset. The prompts are taken from \cite{kirk2024prism}, and any differences in the distribution are due to filtering and difference in topics clustering. }
    \label{fig:topic_distribution}
\end{figure*}

Prior preference datasets \cite{dubois2023alpacafarm, cui2023ultrafeedback} do not have any group or individual-level information. Therefore, in order to empirically study the issues of pluralistic alignment raised earlier, we also construct a wide dataset of preferences based on the population of synthetic personas described in the previous section. We will outline our dataset curation process here.

\textbf{Prompts Curation.} We found the PRISM dataset \cite{kirk2024prism} to contain a diverse set of questions on a multitude of topics, including interpersonal, political, and opinionated issues that can elicit a range of preferences based on the feedback provider's background. To ensure the quality and relevance of the prompts, we performed several post-processing steps. First, we removed any instruction without a question mark and any instruction under five words in length. We then further prompted GPT-4 as a zero-shot classifier to assess whether a question is controversial or not and removed prompts which would not induce diverse opinions. This resulted in a final set of 3868 of the 8011 in the original dataset kept in our final version. The distribution of the discussion topics that are covered in our datasets is shown in Fig. \ref{fig:topic_distribution}. In order to be able to evaluate generalization we split the dataset in 3000 train prompts and 868 held-out prompts which uniformly cover the distribution of topics.

\textbf{Preference Dataset Curation.} While classical RLHF pipelines \cite{stiennon2020learning, ouyang2022training, bai2022constitutional} sample multiple answers from the reference model and asking users to rank those, this procedure is not directly applicable to our setting for several reasons. First, we base all our data generation on synthetic role-playing models, and the quality and instruction-following capabilities of the role-playing model significantly affect the fidelity of answers and feedback. However, all strong openly-available models have already undergone significant RLHF-tuning. As discussed in our introduction and related works, frontier models may have limited diversity in their responses and not fully represent the plurality of views in a demographic. Therefore, to construct a diverse set of preferences, we followed a different approach: We first randomly sample a prompt $x_i$ and a persona $p_i$ in an independent manner. Unlike the PRISM dataset this makes the user profiles independent from the conversational topics. This is a deliberate design choice as directly matching the joint distribution of demographic characteristic and topics in the data could yield models with superficial alignment that learn to map certain topics to the demographic which engages the topic the most and align with those opinions. Instead, we would like to be able to evaluate the whole distribution of opinions and potentially teach the model to elicit preferences and information from the user and not rely on spurious correlations. 

The original PRISM dataset solicits feedback on generations from several models of different sizes and capabilities. Instead we only use GPT 4 for generating answers and as an evaluator for two main reasons; first we want to disentangle the effect of model capability from the model-user alignment and GPT-4 has shown strong role-playing capability. Second, in order to create an easily accessible and reproducible test environment we want to evaluate aligned models under the same preference distribution, which generated the data, hence following prior work \cite{zheng2023judging,dubois2023alpacafarm} in the "LM-as-a-judge" framework, we use also GPT 4 as an evaluator.

\begin{figure*}
    \centering
    \includegraphics[scale=0.25]{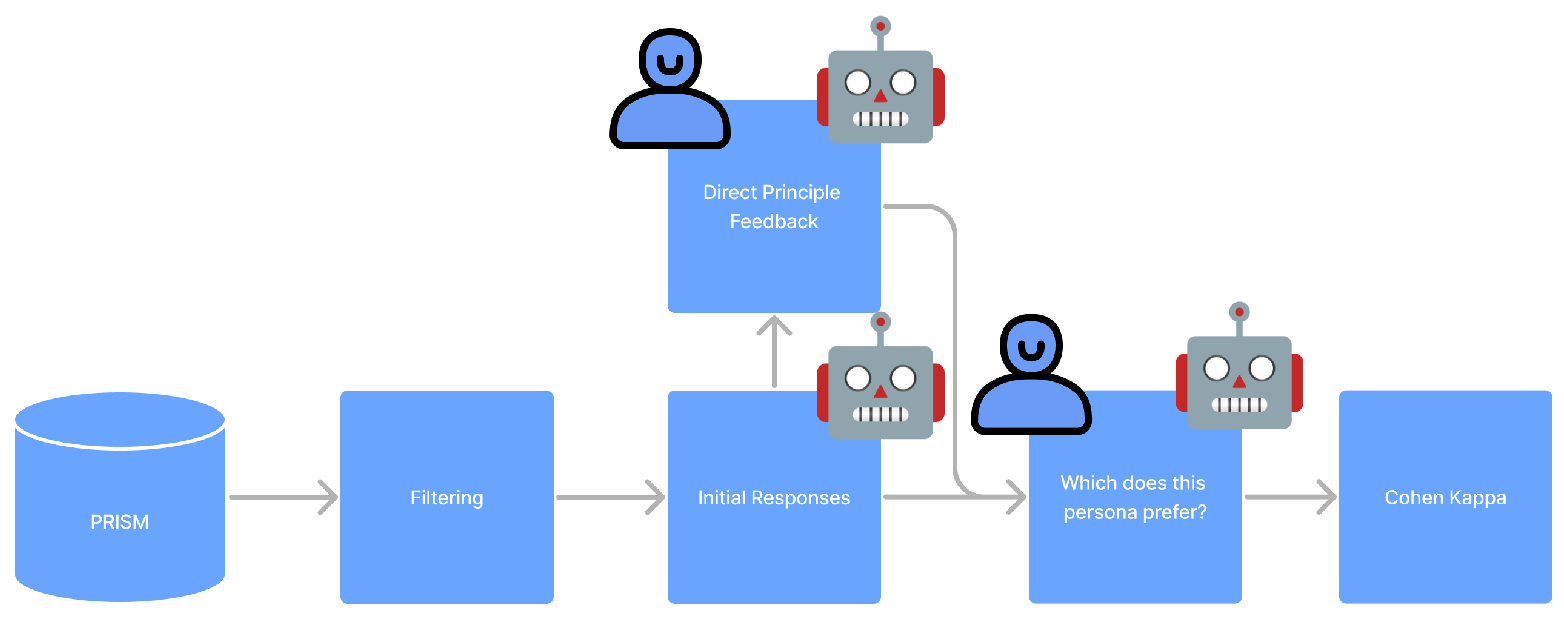}
    \caption{High level for going from the original PRISM dataset to a confusion matrix of Cohen's Kappa between simulated personas. The robot emoji signifies the inclusion of a language model, where as the person emoji signifies the use of a persona (or multiple.)}
    \label{fig:feedback_construction}
\end{figure*}

We construct feedback data using the the Direct Principle Feedback (DPF) approach \cite{castricato2024suppressing} as it tends to outperform Constitutional AI methods \cite{bai2022constitutional}. Our data pipeline is shown in Fig. \ref{fig:feedback_construction}. Once we have the pair of prompts and personas $x_i, p_i$, we sample a response $y_i^l\sim \pi(y|x_i)$ from GPT 4 using only the question and not the providing access to the person profile, which we consider a proxy for the "representative" user. Then, following \cite{castricato2024suppressing} we further provide the initial response and the user profile and ask the model to re-write the response in order to reflect the user's values $y_i^w\sim \pi(y|y_i^l, x_i, p_i, r)$, where $r$ is the DPF query prompt as shown in Appendix \ref{sec:DPF}. We then have the feedback tuple $p_i, x_i, y_i^w\succ y_i^l$ where we assume the persona $p_i$ would always prefer the re-written response over the base model response. When we evaluate the two choices, using a role-playing evaluator, this assumption holds \textbf{96\%} of the time. For every persona we sample 150 prompts from the 3000 train prompts and create a single preference pair per prompt. For personalization and preference elicitation applications, we split the 150 pairs into 100 train prompts and 50 held-out test prompts. We further sample 50 prompts from the 868 held-out test prompts and create an additional 50 preference pairs. In total the dataset contains 100 train preference pairs for each persona and 100 test preference pairs split in 50 seen prompts and 50 held-out prompts for a total of 158,600 total train preference pairs and the same amount of held-out data.







\section{Dataset Analysis and Human Verification}

In this section, we present an analysis of our dataset and the human verification process employed to validate the relevance of persona attributes in the decision-making process.

\subsection{Leave One Out Analysis}

To determine the relevance of persona attributes to the evaluation process, we performed a leave-one-out analysis. For each attribute $a_i$, we randomly constructed 40 personas, each consisting of 3 attributes excluding $a_i$. We then created a corresponding set of 40 personas identical to the first set but with the addition of the LOO attribute $a_i$, for a total of 4 attributes per persona. Our attribute filtering process may have introduced some sampling bias. For example, when analyzing the ``disability type'' attribute, we first filtered our dataset to only include personas with a disability before adding the specific ``disability type'' attribute.


Analogous to conventional leave one out analysis, for every attribute, $a_i$, we had a set of personas without that specific attribute and an analogous set of personas that were identical except for the inclusion of the leave one out attribute. 

We collated a set of 20 questions and baseline answers, which were used for human evaluation (see Appendix for details). For each persona pair $p_{i,j}$ (Original Persona${i,j}$, Original Persona${i,j}$ + LOO Attribute), where $1 \leq i \leq |\text{attributes}|$ and $1 \leq j \leq 40$. We critiqued and refined all 20 baseline answers to make them more personalized for the given persona. The prompt used for this process can be found in the appendix.

\begin{figure*}
    \centering
    \includegraphics[width=0.975\textwidth]
    {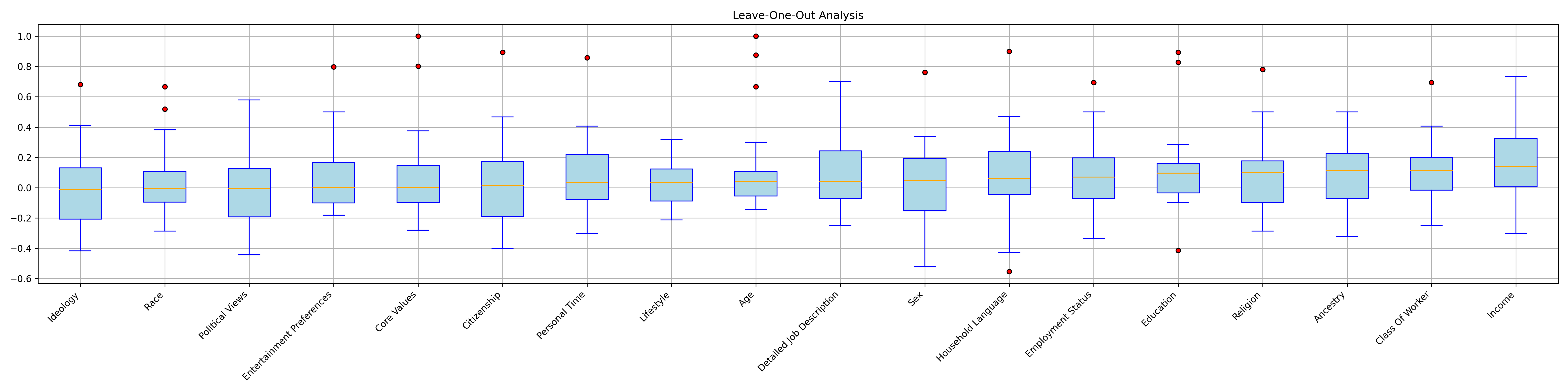}
    \caption{Leave one out analysis of various attributes of our persona. Influence is measured as the annotator agreement (Cohen's kappa) between an annotator with a given attribute and an annotator without said attribute. Lower Cohen's kappa equates to larger influence.}
    \label{fig:loo-analsys}
\end{figure*}

We used Cohen's kappa quantify the agreement between annotators for the original persona and the persona with the LOO attribute concatenated. Cohen's kappa is a statistical measure to assess inter-annotator reliability that takes into account the possibility of agreement occurring by chance. For every pair $p_{i,j}$ we want to measure the annotator agreement between the original persona and the persona with the LOO attribute concatenated. This is repeated $\forall i \text{ s.t. } 1 \leq i \leq |\text{attributes}|, \forall j \text{ s.t. } 1 \leq j \leq 40$. We then report the distributions over these Cohen's kappa per attribute to determine which, if any, attributes are the most influential. The results, as shown in Figure \ref{fig:loo-analsys}, suggest that while the persona as a whole steers the preferences extraction process, no single attribute overpowers the persona.


We've included a number of graphs in the appendix to further explore the relationship between attributes and the overall decision making of personas.

\subsection{Human Evaluation}

Evaluating how humans express preferences is crucial for understanding language models' ability to emulate synthetic personas. Whether humans follow instructions similarly to language models is actively debated \cite{webson2023language}. To validate our approach, we here report inter-annotator agreement between a language model and a human imitating the same persona.
\subsubsection{Experimental Design}
For our human evaluation, we selected 20 personas with a fixed number of attributes, including core values and entertainment preferences. We then recruited 80 participants via Prolific Academic \cite{palan2018prolific}, with each persona shown to 4 independent participants and each rater seeing exactly one persona. We also selected 10 questions for each persona to ``answer'' by initially generating one PRISM refinement step for each persona, starting with 20 questions, and then randomly sampling down to 10 due to human annotation limitations.\footnote{The full set of personas and questions is available here: \href{https://sites.google.com/view/pluralistic}{https://sites.google.com/view/pluralistic}} Each participant was presented with a page outlining what it means to imitate a ``persona'' (see Appendix for instructions). The full annotation UI will be available upon publication. For each persona, we took the majority answer from 3 out of 4 participants.\footnote{The extra annotator allowed for dropping one set of annotations if needed.}






\subsubsection{Results}
Our human evaluation demonstrates that state-of-the-art language models can effectively role-play diverse personas and express preferences aligning with those personas.

\begin{figure}
    \centering
    \includegraphics[scale=0.245]{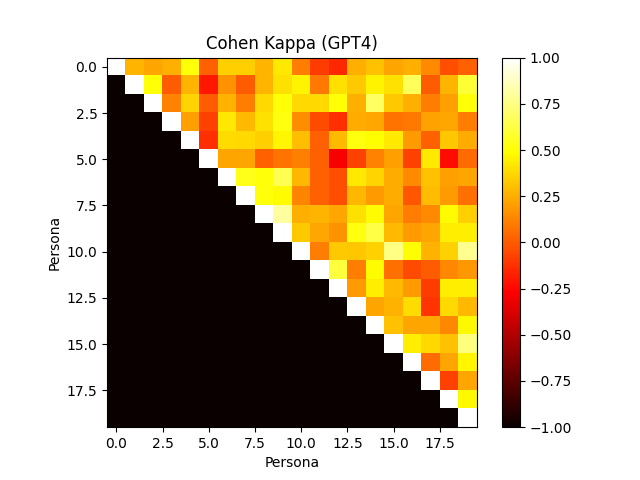}
    \includegraphics[scale=0.245]{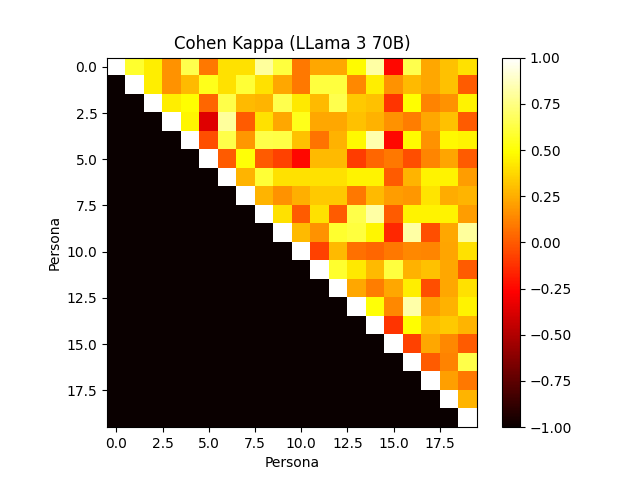}
    \includegraphics[scale=0.245]{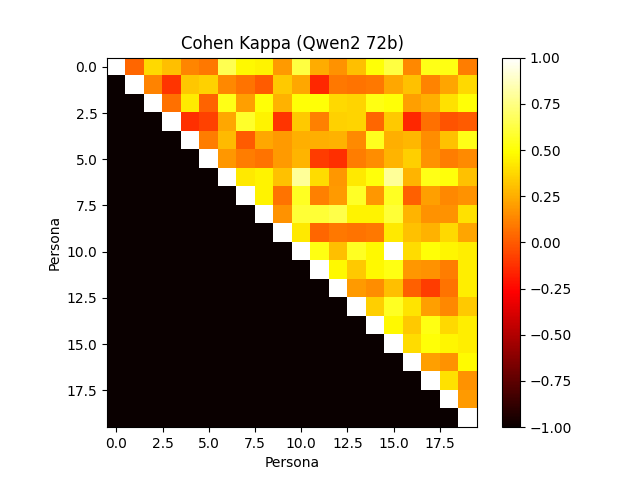}
    \includegraphics[scale=0.245]{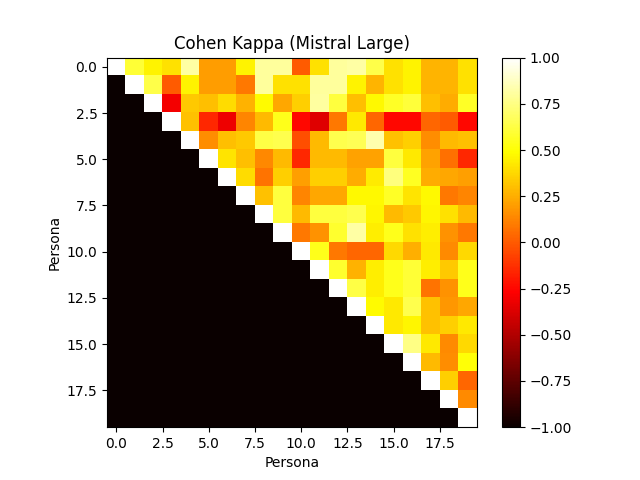}
    \includegraphics[scale=0.7]{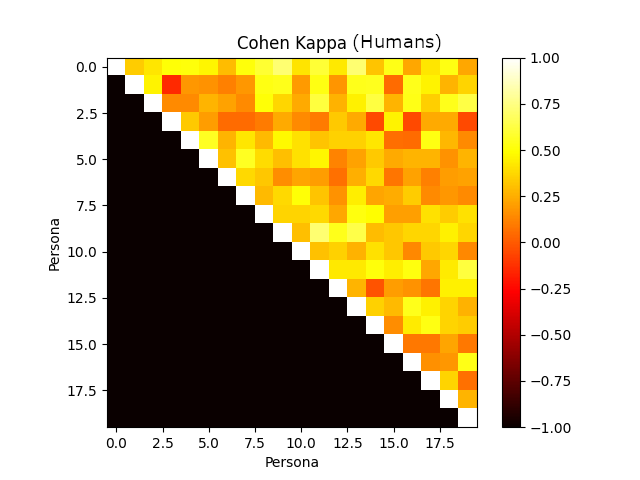}
    \caption{Annotator agreement with various frontier models. Cohen's Kappa confusion matrix. Top left is GPT-4, next is llama 3 70b. Second from the right is Qwen 2 72b. Top right is Mistral Large. Bottom is human vs human inter annotator agreement for a baseline. The lower left triangular matrix is blacked out to keep the scales of the confusion matrices consistent.}
    \label{fig:cohen-kappa-prolific-confusion-matrix}
    \vspace{-5mm}
\end{figure}

\begin{figure}[t]
    \centering
    \includegraphics[scale=0.245]{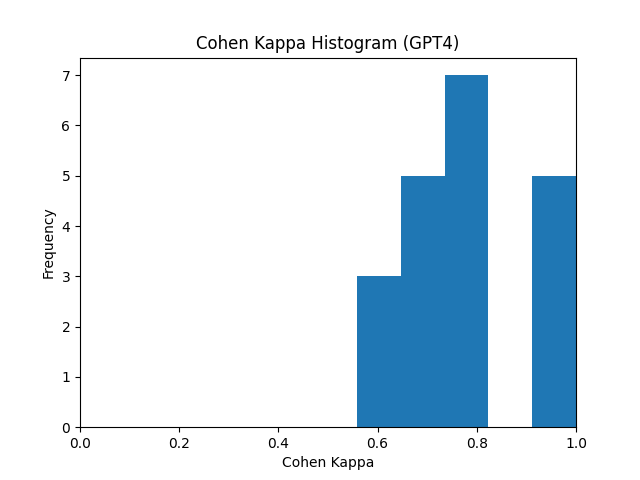}
    \includegraphics[scale=0.245]{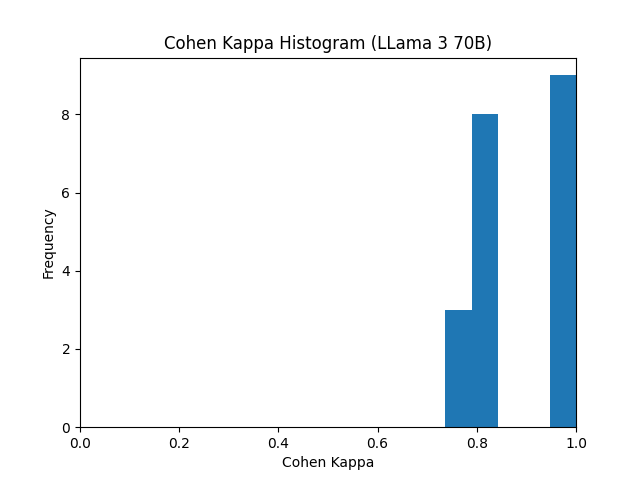}
    \includegraphics[scale=0.245]{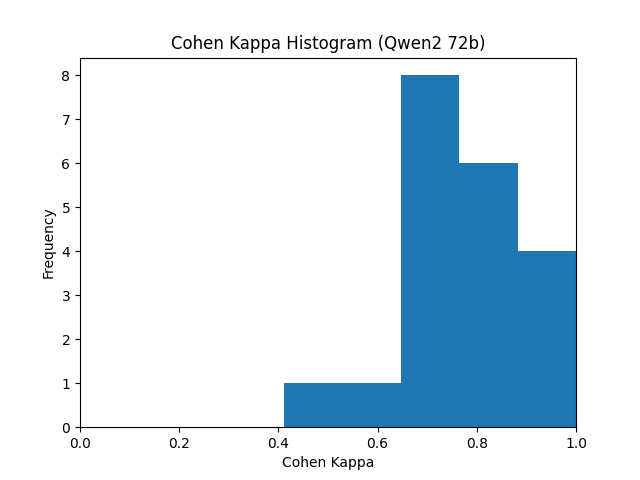}
    \includegraphics[scale=0.245]{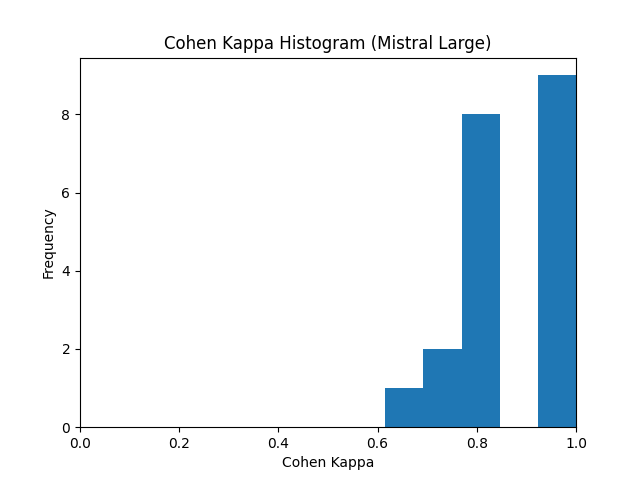}
    \caption{Annotator agreement with various frontier models. Cohen's Kappa histogram. Top left is GPT-4, followed by llama 3 70b top right is LLama-3 70b, Qwen 2 72b, and Mistral. Note that, evident by this graph, Llama 3 70b and Mistral Large have some of the largest annotator agreements, where as GPT-4 and LLama-3 70b have some of the largest spreads of opinions they express, with relatively low accuracy.}
    \label{fig:cohen-kappa-prolific-histo}
    \vspace{-5mm}
\end{figure}

Both figures \ref{fig:cohen-kappa-prolific-confusion-matrix} and \ref{fig:cohen-kappa-prolific-histo} shows the annotator agreement, measured by Cohen's Kappa, between human participants and various frontier language models (GPT-4, LLama-3 70b, Qwen 2 72b, Mistral Large) when imitating the same personas. Notably, GPT-4 achieves high agreement with human annotators, with Kappa values concentrated in the 0.6-0.8 range (substantial agreement). This suggests GPT-4 can accurately capture and express persona-specific preferences in a human-like manner.

However, the persona role-playing capabilities vary across models. As evident in Figure \ref{fig:cohen-kappa-prolific-histo}, Llama-3 70b and Mistral Large exhibit higher annotator agreement compared to GPT-4 and Qwen 2 72b. The latter two models show a wider spread of expressed opinions with lower accuracy. This indicates that while all models can role-play to some extent, their ability to align with human-like persona preferences is not uniform.

\begin{figure}
    \centering
    \includegraphics[scale=0.23]{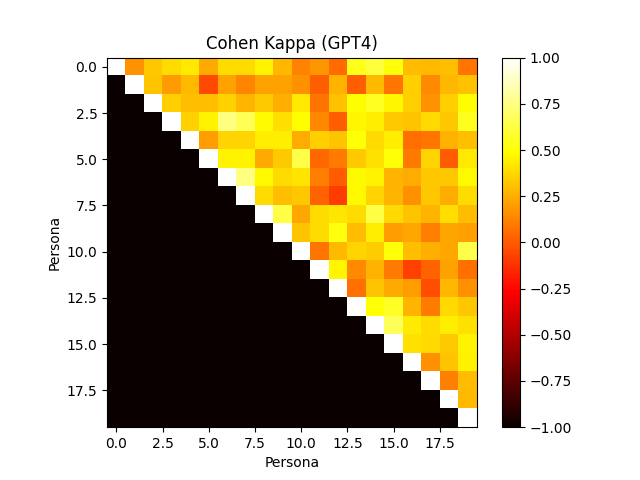}
    \includegraphics[scale=0.23]{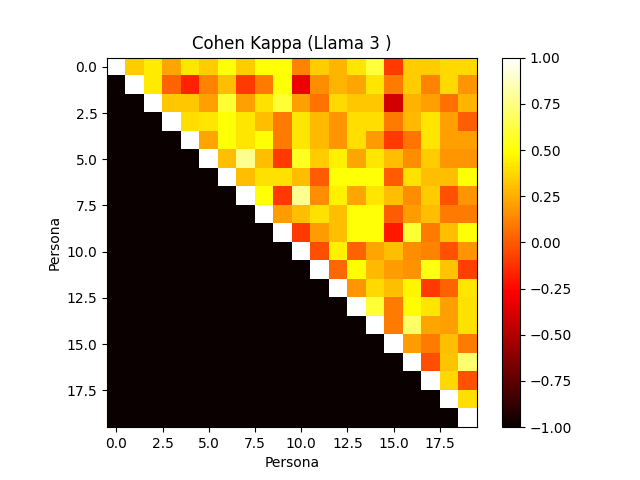}
    \includegraphics[scale=0.23]{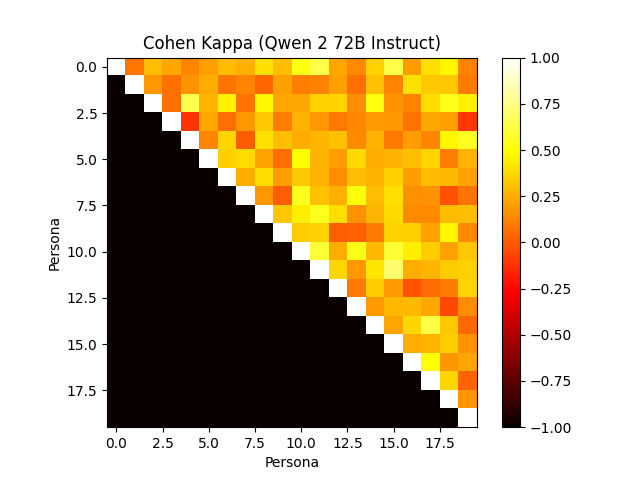}
    \includegraphics[scale=0.23]{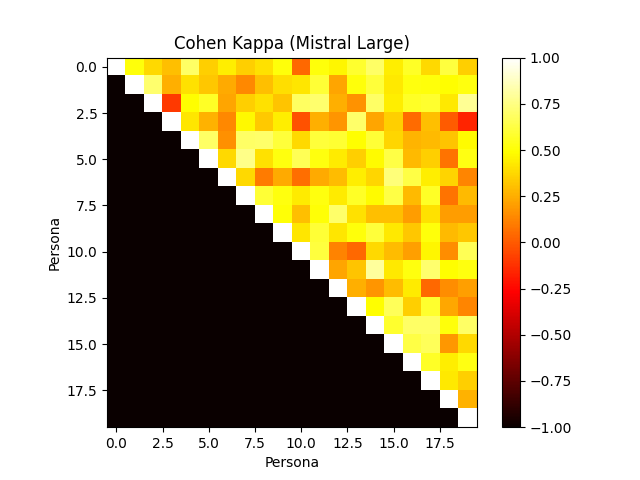}
    \caption{Inter annotator agreement (confusion matrices) for solely frontier model generated persona preferences. Top left is GPT-4, followed by llama 3 70b top right is LLama-3 70b, Qwen 2 72b, and Mistral.}    
    \label{fig:cohen-kappa-confusion}
\end{figure}

\begin{figure}[h]
    \centering
    \includegraphics[scale=0.23]{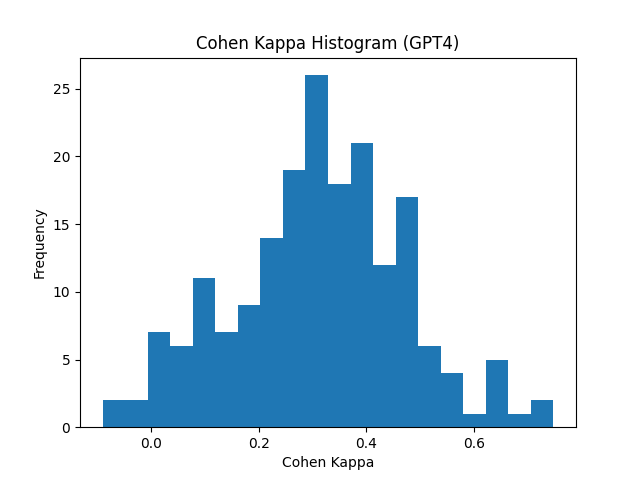}
    \includegraphics[scale=0.23]{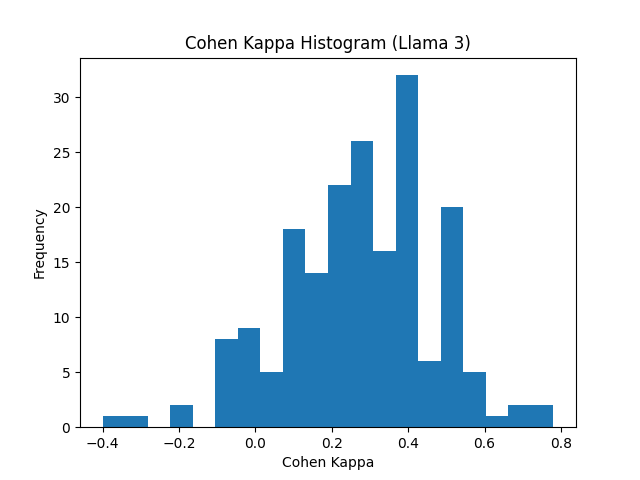}
    \includegraphics[scale=0.23]{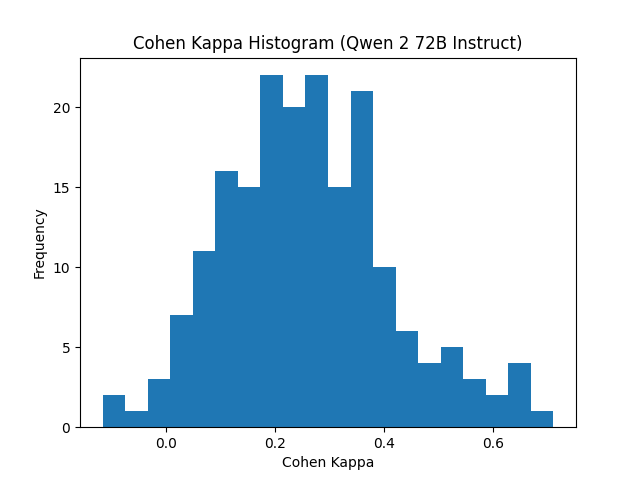}
    \includegraphics[scale=0.23]{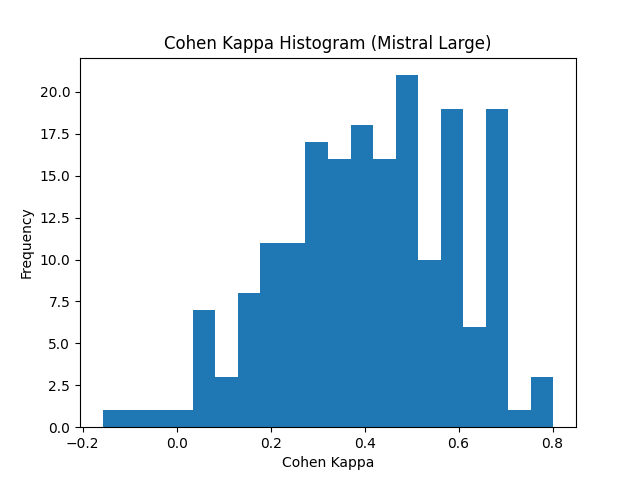}
    \caption{Inter annotator agreement (histograms) for solely frontier model generated persona preferences. Top left is GPT-4, followed by llama 3 70b top right is LLama-3 70b, Qwen 2 72b, and Mistral.}
    \label{fig:cohen-kappa-histo}
\end{figure}

To further investigate the models' role-playing consistency, we examine the inter-annotator agreement between the models themselves when imitating the same personas (Figures \ref{fig:cohen-kappa-confusion} and \ref{fig:cohen-kappa-histo}). The confusion matrices reveal substantial agreement between models, with GPT-4 showing the highest consistency. The histograms confirm this trend, with GPT-4 exhibiting a tight distribution of high Kappa values. 

These results validate our approach of using language models as synthetic personas for evaluating pluralistic alignment techniques. The high agreement between GPT-4 and human annotators, along with the inter-model consistency, suggests that carefully designed role-playing scenarios with language models can serve as a realistic and scalable testbed for assessing alignment methods without the need for human participants.

\


\section{Extensive evaluation of personalization via LLM as a judge}

\begin{figure}[h]
    \centering
    \includegraphics[width=0.9\columnwidth]{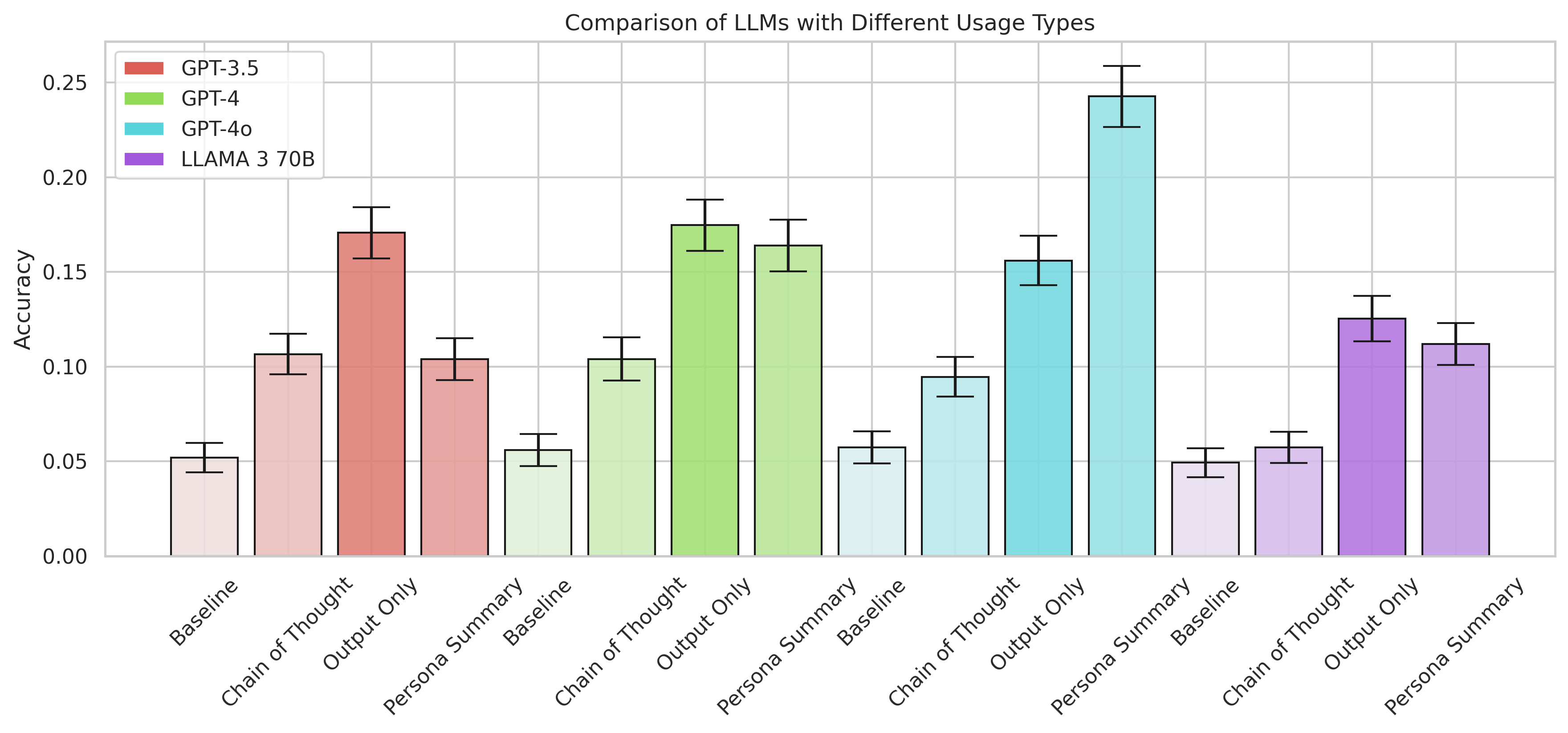}
    \caption{How well various methods of conditioning the PRISM answer generation work. Notice that summarization is by far the best result, with no chain of thought being beaten out by chain of thought. N = 700 questions randomized across 1000 personas.}
    \label{fig:overall-eval}
\end{figure}

Determining a language model's ability to personalize, beyond that of grounding our personas and personalization with human evaluation, can be done via utilizing a critique-in-the-loop mechanism. Namely, we can generate an utterance from a model that is provided the persona and determine if the utterance requires a revision to be correctly subtly personalized. Given the above, we can note on our persona that there is a strong correlation between a language model's ability to imitate a persona and a human annotator's ability to imitate a persona. We've included a number of summary statics about these results in Figure \ref{fig:overall-eval}.

To that end, we present PERSONA Bench, a grounded and human verified pluralistic alignment benchmark that opens the door to extensive discussion around language model's capabilities to align to various personas. \footnote{The full bench marking suite is released as open source under an Apache 2.0 and is available on our GitHub. We include a number of evaluations of interest, inspired by the prior results in this paper.}

The following section is a number of comparison benchmarks we conducted using PERSONA Bench on a number of different generation techniques, varying from not giving the language model being evaluated the persona in question to evaluating the usefulness of chain of thought in the construction of personalized utterances.

\subsection{Baseline Performance}

For a baseline, we wanted to measure the performance of a model's ability to personalize for a given persona when it is not informed of any details about this persona and not allowed to generate chain of thought. Obviously, one would expect lackluster results in this case as being unable to know what demographic of personas you are supposed to personalize for would result in significant complications with downstream performance. These effects can be directly observed in the quantitative performance of this baseline, averaging approximately 5\% accuracy across the board almost uncorrelated of model size and performance.

\subsection{Efficacy of adding chain of thought to personalized answer generation}

One of the effects that we observed is that the inclusion of chain of thought actually degraded performance, often to quite detrimental levels. We refer the reader to Figure \ref{fig:overall-eval}.

We originally hypothesized that this was in part due to the language model in question utilizing chain of thought as a vector to regurgitate various aspects about the persona, therefore resulting in a penalty from the critic as the response being non-subtle, but we found that across 3000 question/answer pairs the chain of thought answers where negligibly more likely to be non-subtle, about a 3\% increase with little to no statistical significance (33\% vs 36\%). This was done via LLM as a judge, utilizing GPT-4o.  

\subsection{Summarization of Persona}

Branching off some of the results we saw in the chain of thought work, we hypothesized that if we provided chain of thought with more guidance, via asked the language model to first summarize the parts of the persona that were inherently relevant to the PRISM question before answering the question itself, then we would observe a significant performance increase. This bears a resemblance to Lost in the Middle, \cite{liu2023lostmiddlelanguagemodels}, where the authors observe a sizeable performance increase by first extracting relevant information before answering a question.

This approach resulted in the most performant evaluation, accross all models that we tested, as shown in Figure \ref{fig:overall-eval}. 

This, however, rose the question about the kinds of transformations that this summarization step was doing. To evaluate this, we devised an LLM-as-a-judge procedure where we asked GPT-4o-mini was tasked with finding textual differences between the persona summary compared to the original persona.

As a sanity check, the LLM-as-a-judge reported that none of the personas and persona summaries were equivalent with respect to the question being asked.

\begin{figure*}
    \centering
    \includegraphics[width=\textwidth]{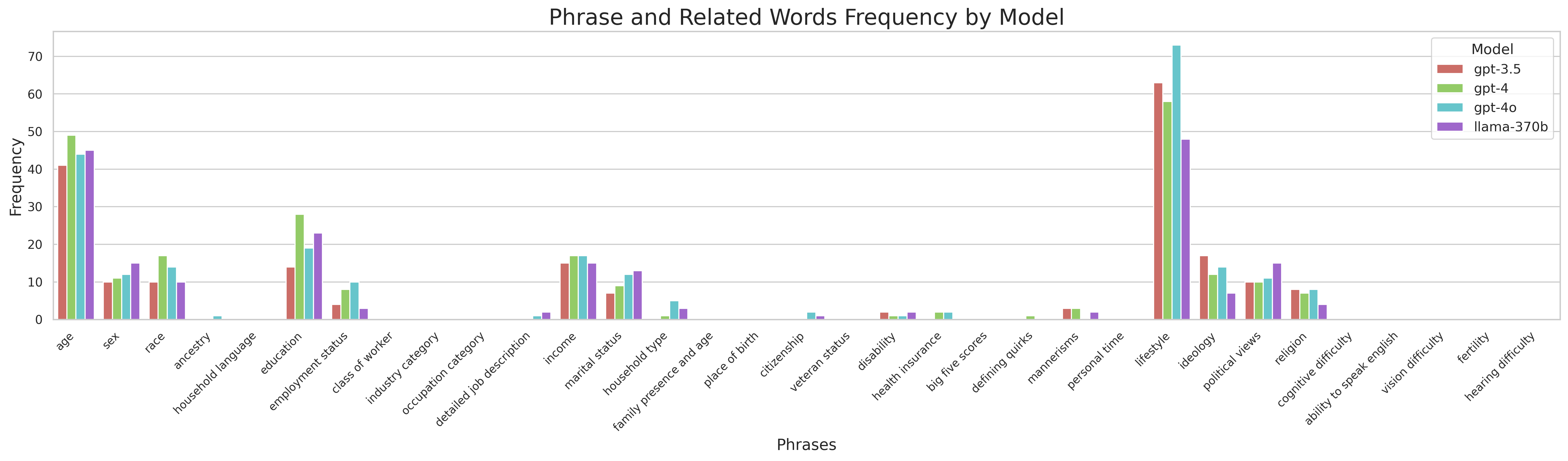}
    \caption{This figure highlights the frequency that the summarization step would remove various persona related attributes when producing the summary. Notice that factors like age and lifestyle choice were the most often ones to be removed. The attributes removed are pretty consistent across models.}
    \label{fig:dropped-phrases}
\end{figure*}

We measured the frequency that the difference that GPT-4o-mini produced contained key persona attributes, or synonyms of those attributes, to determine how often the summarization step was removing certain components of the persona from a near context window. The results from this frequency analysis can be found in Figure \ref{fig:dropped-phrases}. We found that by far factors like age and lifestyle were most likely to be removed during the summarization step, as perhaps those were the least relevant to the variety of questions found in the PRISM dataset. Furthermore, we found that the attributes removed were incredibly consistent across models and even across models on the same question. Further work is needed here to fully understand the mechanisms and biases at play.

\subsection{Pass@K Evaluation}

One of the primary challenges in determining which model to deploy in production is a model's scalability with respect to compound inference. To that end, we ran our main evaluation (chain of thought + providing the LLM with the persona in question) on pass@k for k up to 16. We note that some of the models that are traditionally very performant and safe models, like GPT-4o and GPT-3.5, are actually some of the worst performing where as Llama 3 70b manages to overtake all tested OpenAI models.

\begin{figure}
    \centering
    \includegraphics[width=0.8\textwidth]{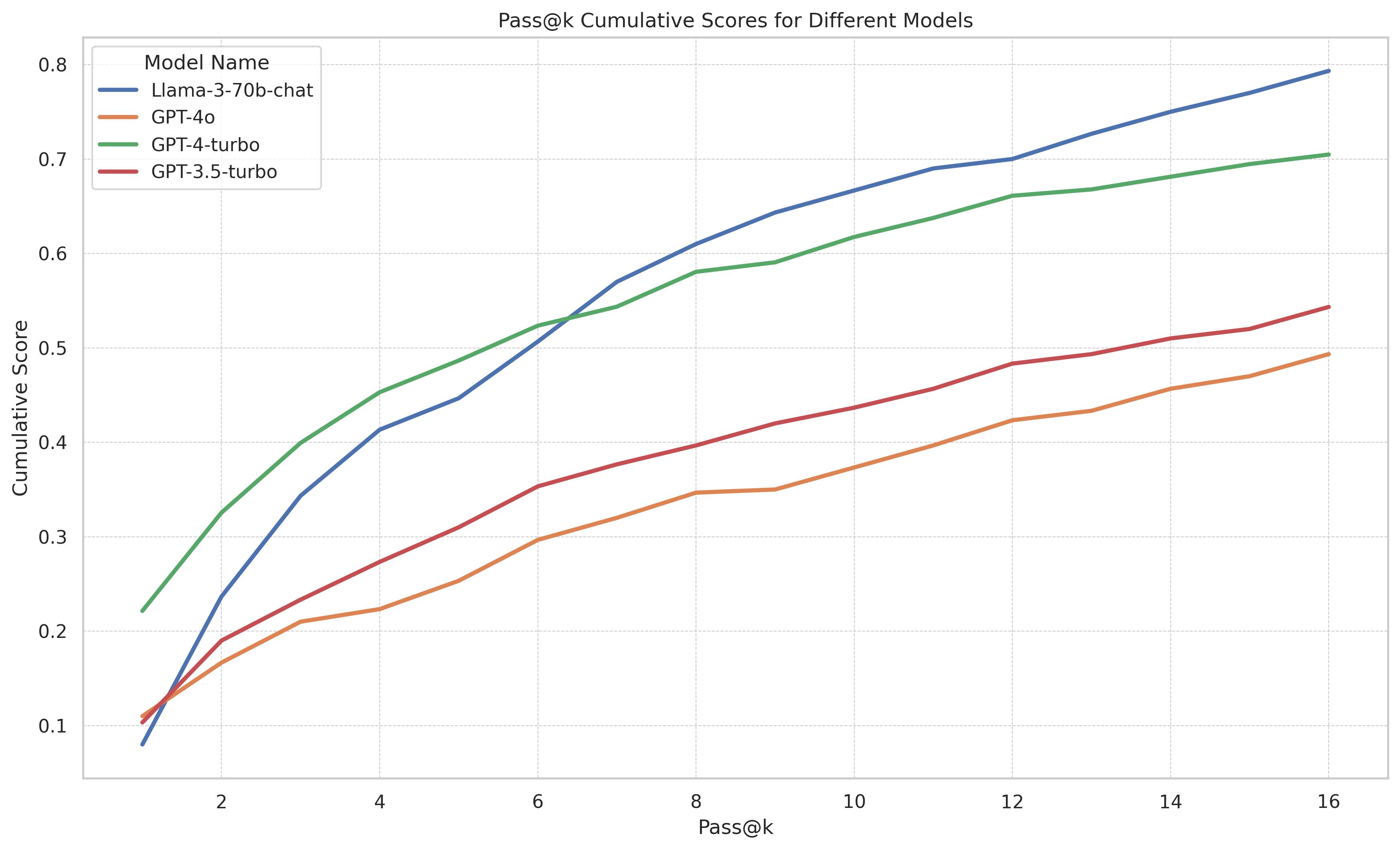}
    \caption{Pass@16 results for various models on the main evaluation, namely chain of thought plus providing the large language model with the same persona that the LLM-as-a-judge will be provided.}
    \label{fig:enter-label}
\end{figure}

\section{Conclusion}

The advancement and wide adoption of language models has raised a number of important concerns around fairness and pluralistic alignment to the values of diverse users, which still remains a challenge. Beyond group-level preferences, personalized models, tailored to specific individual needs and preferences are a promising application. Despite the concerns and opportunities raised by these issues, current large-scale RLHF pipelines still work under the assumption of a representative user and do not account for the distributional nature of values. While a number of academic works have proposed approaches for pluralistic alignment, personalization and preference elicitation, these are still not widely adopted, partially due to lack of convincing evaluations as current benchmark consists of unrealistic multiple-choice questions or simple domains. 
In this work we aim to address this challenge by creating a test environment and benchmark for these issues. We propose an automated LM as-a-judge approach based on current state-of-the-art systems role-playing capabilities. We create a demographic of 1000 train and 568 test realistic \texttt{personas} based on US census demographics and individualized profiles with idiosyncratic personality types. We further utilize a wide real user survey controversial topics to create a large-scale synthetic datasets of diverse feedback with over 158,600 train preference pairs and a comparable number of evaluation datapoints. Our proposed environment can be used to develop and evaluate approaches on pluralistic alignment with diverse group preferences, individualized models and information-gathering and preference elicitation. We further validate the fidelity of these personas with a real user study.

We believe our work will facilitate the developemnt of new alignment approaches, but a open questions remain. In this construction we focused exclusively on US demographics and user profiles, which are not representative of global populations. These users might already be over-represented in LM training data (hence the advanced role-playing capabilities of GPT 4 on this demographic). 

Further work would evaluate different LM model's capabilities to represent a global audience and expand the persona demographics to include these populations as as well.

\clearpage
\pagebreak
\section{Limitations}
Our work has several potential limitations.


\textbf{Demographic Focus:} Our personas are based on US demographic data, which may not accurately represent the diversity of global populations. This limitation could impact the generalizability of our findings to non-US contexts. Future work should aim to include a more diverse set of personas reflecting global demographic and cultural variations.

\textbf{Feedback and Preference Data:} The preference data generated in this study relies on the responses of language models in role-playing scenarios. While we validated these responses through human judges, there remains a risk that the feedback does not perfectly mimic real human preferences. Additionally, the Direct Principle Feedback (DPF) approach, although effective, may not capture all nuances of human decision-making and preference.

\textbf{Model Limitations:} The language models used to generate and evaluate personas are themselves subject to biases and limitations. Current state-of-the-art models, such as GPT-4, have shown strong role-playing capabilities, but they are not infallible and may produce outputs that are biased or inconsistent. Moreover, the role-playing capabilities of these models might not extend uniformly across different types of personas, especially those representing underrepresented or marginalized groups.

\textbf{Evaluation Metrics:} The use of Cohen’s kappa and other inter-annotator agreement metrics provides a measure of consistency but may not fully capture the qualitative aspects of alignment with human preferences. These metrics focus on agreement rates, which do not necessarily reflect the richness and contextual appropriateness of the model’s responses.

\textbf{Real-World Application:} While our synthetic approach allows for scalable testing and evaluation, it does not fully address the challenges of real-world deployment. The dynamics of real user interactions, continuous learning, and adaptation to evolving preferences are complex and require more extensive field testing and longitudinal studies.

\textbf{Bias Concerns:} The creation and use of synthetic personas must be approached with caution to avoid perpetuating stereotypes or introducing new biases. Our study attempts to mitigate these risks through careful design and validation, but there remains a possibility that some biases are not fully addressed.

In summary, while PERSONA provides a valuable testbed for evaluating pluralistic alignment in language models, these limitations highlight the need for ongoing research and development to refine these methods and ensure their applicability and fairness in diverse real-world settings.

\clearpage
\pagebreak

\bibliography{main_latex}

\appendix

\clearpage
\pagebreak

\section{Full list of attributes}\label{sec:persona_attributes}

The following is the full list of persona attributes.

\begin{enumerate}
  \item age
  \item sex
  \item race
  \item ancestry
  \item household language
  \item education
  \item employment status
  \item class of worker
  \item industry category
  \item occupation category
  \item detailed job description
  \item income
  \item marital status
  \item household type
  \item family presence and age
  \item place of birth
  \item citizenship
  \item veteran status
  \item disability
  \item health insurance
  \item big five scores
  \item defining quirks
  \item mannerisms
  \item personal time
  \item lifestyle
  \item ideology
  \item political views
  \item religion
  \item cognitive difficulty
  \item ability to speak English
  \item vision difficulty
  \item fertility
  \item hearing difficulty
\end{enumerate}

\clearpage
\pagebreak

\section{Persona Critique and Refinement Prompt}\label{sec:DPF}

The following is the critique prompt that was used.

\begin{verbatim}
f"Examine the COMPLETION: '{preference}' in relation "
"to the DEMOGRAPHIC: '{persona}' and the INSTRUCTION: " '{preference.meta_data['instruction']}'. "
"Put yourself in the shoes of DEMOGRAPHIC. "
"The demographic prefers short answers.  "
" If you give a long suggestion, they will hate it. "
"Identify the ways the completion both does and does not resonate with the demographic. "
"Provide a concise explanation, quoting directly from the demographic 
and completion to illustrate your evaluation. "
"Think step by step about how you will make the response shorter or the same length before 

providing your evaluation and suggestions. "
"Similarly, make sure that the response given is still relevant to the INSTRUCTION. "
"Format: EVALUATION: ... SUGGESTIONS: ...\nDONE"
\end{verbatim}

The following is the revision prompt that was used.

\begin{verbatim}
f"Revise the COMPLETION: '{preference}', "
"with respect to INSTRUCTION: " "'{preference.meta_data['instruction']}' 

based on the CRITIQUE: '{critique}'. "
"Provide a revision of the completion, do not make ANY "
"references to the exact preferences or attributes "
"of the demographic. "
f"Remain subtle and indirect in your revision. "
"Make sure your response has less tokens than the original completion. "
"If you make it longer you are a BAD CHATGPT. "
"Format: REVISED PREFERENCE: ...\nDONE"
\end{verbatim}

\clearpage
\pagebreak

\section{Complete Example Persona}\label{sec:example_personas}

The following is an example of a persona

\begin{verbatim}

 'age': 73,
 'ancestry': 'Filipino',
 'big five scores': 'Openness: Extremely High, Conscientiousness: Low, '
                    'Extraversion: Extremely High, Agreeableness: Low, '
                    'Neuroticism: Extremely Low',
 'citizenship': 'U.S. citizen by naturalization',
 'class of worker': 'Retired',
 'cognitive difficulty': nan,
 'defining quirks': 'Enjoys gardening and has a green thumb',
 'detailed job description': 'Retired, previously worked in a managerial '
                             'position',
 'disability': nan,
 'education': "Bachelor's Degree",
 'employment status': 'Not in labor force',
 'family presence and age': 'With related children 5 to 17 years only',
 'fertility': nan,
 'health insurance': 'With health insurance coverage',
 'hearing difficulty': nan,
 'household language': 'Asian and Pacific Island languages',
 'household type': 'Married couple household, no children of the householder '
                   'less than 18',
 'ideology': 'Liberal',
 'income': '178900',
 'industry category': nan,
 'lifestyle': 'Active and outdoorsy',
 'mannerisms': 'Often uses hand gestures while speaking',
 'marital status': 'Married',
 'occupation category': nan,
 'personal time': 'Spends free time gardening or reading',
 'place of birth': 'Philippines',
 'political views': 'Democrat',
 'race': 'Asian',
 'religion': 'Other Christian',
 'sex': 'Female',
 'veteran status': 'Non-Veteran',
 'vision difficulty': nan}
 
\end{verbatim}

\clearpage
\pagebreak

\begin{verbatim}

 'ability to speak english': nan,
 'age': 10,
 'ancestry': 'Mixed',
 'big five scores': 'Openness: Extremely High, Conscientiousness: Average, '
                    'Extraversion: Extremely Low, Agreeableness: Extremely '
                    'High, Neuroticism: Average',
 'citizenship': 'Born in the United States',
 'class of worker': 'Not applicable',
 'cognitive difficulty': nan,
 'defining quirks': 'Prefers to express herself through drawing',
 'detailed job description': 'Student',
 'disability': nan,
 'education': 'Grade 3',
 'employment status': 'Unemployed',
 'family presence and age': 'With related children under 5 years and 5 to 17 '
                            'years',
 'fertility': nan,
 'health insurance': 'With health insurance coverage',
 'hearing difficulty': nan,
 'household language': 'Spanish',
 'household type': 'Married couple household with children of the householder '
                   'less than 18',
 'ideology': 'Believes in fairness and kindness',
 'income': '0',
 'industry category': 'Not applicable',
 'lifestyle': 'Active and curious',
 'mannerisms': 'Often hums while concentrating',
 'marital status': 'Never married or under 15 years old',
 'occupation category': 'Student',
 'personal time': 'Spends free time drawing or reading',
 'place of birth': 'California/CA',
 'political views': 'Too young to have political views',
 'race': 'Two or More Races',
 'religion': 'Protestant',
 'sex': 'Female',
 'veteran status': 'Not applicable',
 'vision difficulty': nan}
\end{verbatim}

\clearpage
\pagebreak

\section{Annotation Instructions}

Welcome to the Persona Annotation Task!<br><br>
In this task, you will be asked to role-play as a specific persona and answer a series of preference questions. <br>
<strong>1. Task Explanation:</strong> We will provide you with a set of descriptors of a particular person. This person may or may not actually exist. Your job is to put yourself into the mindset of a person with those attributes.<br>
<strong>2. Instruction following:</strong> You will be presented with a hypothetical question that a person could ask. Your job is to select the answer that a person with the attributes that you are impersonating would prefer. <br>
<strong>3. Explain your reasoning:</strong> Justify your choice. It is ok to change your choice while thinking through your justifcation. In the textbox provided below the prefernece selection, go into detail about why you think your choice is correct. If there is no clear choice, pick the one that is most likely, just still attempt to justify your selection.<br>
<strong>4. Provide good reasoning:</strong> The better your reasoning, the bigger your <strong>bonus</strong> will be.<br>
<strong>5. ChatGPT (or other chatbots) are NOT allowed:</strong> Any use of ChatGPT for soliciting preferences or reasoning will result in disqualification. <br>
You <strong>must</strong> each question based on how you think the given <strong>persona</strong> would respond, not based on your personal preferences. <br><br>
Thank you for participating!

\clearpage
\pagebreak

\section{Census evaluation}

\begin{figure*}
    \centering
\includegraphics[width=0.975\textwidth]{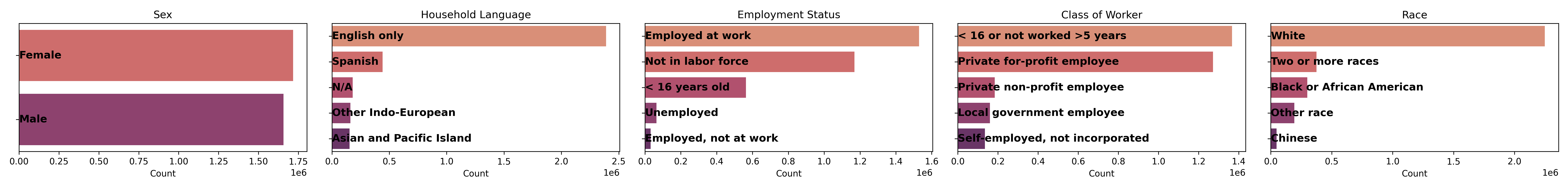}
    \caption{Histogram of demographics statistics from US Census \cite{census_bureau_acs_microdata}.}
    \label{fig:real census}
\end{figure*}

One of the most important aspects of our evaluation in Figure \ref{fig:persona_demographics} was that our personas correlated strongly with a US census data baseline. Similar to Figure \ref{fig:persona_demographics}, we collated the same data from real US census data to show a comparison to a baseline. We find that our results correlate strongly if not perfectly with ground truth data. The equivalent graph can be found in Figure \ref{fig:real census}.

\end{document}